\documentclass[11pt,twocolumn]{article}

\usepackage[utf8]{inputenc}
\usepackage[T1]{fontenc}
\usepackage{times}
\usepackage{latexsym}
\usepackage{amsmath,amssymb}
\usepackage{graphicx}
\usepackage{booktabs}
\usepackage{multirow}
\usepackage{hyperref}
\usepackage{url}
\usepackage{natbib}
\usepackage{xcolor}
\usepackage{tikz}
\usepackage{pgfplots}
\usepackage{subcaption}
\usepackage[margin=1in]{geometry}
\usepackage{algorithm}
\usepackage{algpseudocode}
\usepackage{xspace}

\pgfplotsset{compat=1.17}
\usetikzlibrary{shapes,arrows.meta,positioning,fit,calc}

\newcommand{\ours}{\textsc{Prove}\xspace}
\newcommand{\oursexpanded}{\textsc{Prove} (Programmatic Rewards On Verified Environments)\xspace}
\newcommand{\rvalidity}{R_{\text{validity}}}
\newcommand{\rcoverage}{R_{\text{coverage}}}
\newcommand{\refficiency}{R_{\text{efficiency}}}
\newcommand{\rtotal}{R_{\text{total}}}

\title{Synthesize and Reward --- Reinforcement Learning for Multi-Step Tool Use in Live Environments}

\author{
  {\large Ibrahim Abdelaziz\thanks{Correspondence: \texttt{ibrahim.abdelaziz1@ibm.com}}, Asim Munawar, Kinjal Basu, Maxwell Crouse,} \\
  {\large Chulaka Gunasekara, Suneet Katrekar, Pavan Kapanipathi} \\
  \textit{IBM Research}
}

\date{}

\begin{document}
\maketitle

\begin{abstract}
Training LLMs to orchestrate multi-step tool calls is held back by three coupled obstacles: realistic stateful execution environments are costly to build, synthetic training queries are often detached from the server's actual state (so the generated tool calls fail to execute), and recall-based RL rewards incentivize verbose tool-calling patterns.
We present \oursexpanded, a framework with three contributions:
(1)~a library of 20 stateful MCP (Model Context Protocol) servers exposing 343 tools, enabling live-execution RL training with session-scoped state isolation; 
(2)~a state-machine data synthesis pipeline that generates multi-turn tool-call trajectories \emph{grounded in live-sampled server state}, so generated queries reference entities that actually exist;
and (3)~a multi-component programmatic reward with an adaptive efficiency penalty that counters the verbosity incentive of recall-based rewards.
We train four models (Qwen3-4B, Qwen3-8B, Qwen2.5-7B, Granite-4.1-8B) with GRPO on the resulting $\sim$13K training examples.
On BFCL Multi-Turn, $\tau^2$-bench, and T-Eval, \ours yields improvements of up to $+10.2$, $+6.8$, and $+6.5$ points respectively, demonstrating that this framework yields consistent gains on multi-step tool orchestration across two model families.
\end{abstract}

\section{Introduction}
\label{sec:intro}

Training LLMs to orchestrate multi-step tool calls---selecting the correct API, filling parameters from context, and respecting inter-call data dependencies---is a prerequisite for deploying autonomous agents in production \citep{yao2023react, schick2023toolformer}.
The problem is harder than single-turn function calling: the model must propagate intermediate results, handle missing information gracefully, and abstain when no tool is applicable.
Two complementary lines of work have emerged in response.

\emph{Data synthesis and supervised fine-tuning.}
xLAM \citep{zhang2024xlam} releases a large corpus of verified single-turn function-calling traces, and APIGen-MT \citep{liu2025apigenmt} extends this to multi-turn trajectories.
More recent pipelines target realism and scale for multi-turn tool use: Magnet \citep{yin2025magnet} composes traces from tool-dependency graphs, ToolACE-MT \citep{zeng2026toolacemt} introduces non-autoregressive multi-turn generation, and TOUCAN \citep{xu2025toucan} scales synthesis to 1.5M traces against real MCP servers.
These pipelines typically feed a separate SFT stage and are not coupled to execution feedback during training.

\emph{Reinforcement learning for tool use.}
A parallel line uses RL to shape tool-use behavior from reward signals without large SFT corpora.
ToolRL \citep{chen2025toolrl}, ReTool \citep{feng2025retool}, StepTool \citep{yu2024steptool}, Tool-R1 \citep{zhang2025toolr1}, Nemotron-Research-Tool-N1 \citep{zhang2025tooln1}, MUA-RL \citep{zhao2025muarl}, and Tool Zero \citep{zeng2025toolzero} explore per-call and step-grained rewards, multi-turn user simulation, and RL-from-scratch.
Closest to our setting, \citet{cheng2026toolorch} decomposes the reward into per-call validity and sequence-level consistency on a single cached-API domain, while the concurrent AgenticQwen \citep{agenticqwen2026} scales to $\sim$100K branching trajectories with LLM-as-judge rewards (Qwen3-235B).

Despite this progress, three gaps remain unaddressed jointly:
(i) most pipelines execute against cached or simulated APIs, missing the state-dependent failure modes that arise in live, stateful services;
(ii) synthetic queries are often detached from the server's actual state, producing traces whose parameters do not correspond to any real entity;
and (iii) sequence-level rewards based purely on recall incentivize \emph{verbosity}---RL-trained models emit excess tool calls to maximize GT coverage, a regression that is rarely penalized explicitly.

We close these gaps with \oursexpanded,\footnote{Code, the 20 MCP server library (343 tools), synthesized training data, and trained model checkpoints will be released soon.} a framework that tightly couples data synthesis to the RL loop through a shared live-execution backend:
\begin{enumerate}
    \item \textbf{Live MCP environments.} A library of 20 stateful Model Context Protocol servers exposing 343 tools, with session-scoped state isolation, used for both data synthesis and RL training. Unlike cached APIs, these capture realistic state-dependent execution dynamics.
    \item \textbf{Grounded state-machine data synthesis.} A pipeline that probes each live server for real entities, grounds query generation in that sampled state, drives a state-machine teacher through multi-turn conversations against the server, and replay-validates every trace. This produces $\sim$13K training examples (multi-turn MCP conversations, missing-function clarification, and abstention) without manual annotation or an LLM-as-judge.
    \item \textbf{Programmatic multi-component reward.} A five-part reward decomposing tool-use quality into validity, dependency-ordered coverage, an \emph{adaptive efficiency penalty} with a complexity-scaled call budget, a tool-name signal, and an argument-value matching bonus---fully programmatic with no external judge model (\S\ref{sec:reward}). An ablation (\S\ref{sec:reward_ablation}) shows that the adaptive budget and the tool-name signal are the two most load-bearing components.
\end{enumerate}
Table~\ref{tab:summary_deltas} summarizes results across four models from two architecture families, trained with identical hyperparameters.
\begin{table*}[t]
\centering
\caption{Improvement over base model ($\Delta$) after training with \ours{}. All models trained for 350 GRPO steps.}
\label{tab:summary_deltas}
\resizebox{0.55\textwidth}{!}{%
\begin{tabular}{l|cc|cc|cc}
\toprule
\textbf{Model} & \textbf{BFCL MT} & $\boldsymbol{\Delta}$ & \textbf{TAU2} & $\boldsymbol{\Delta}$ & \textbf{T-Eval} & $\boldsymbol{\Delta}$ \\
\midrule
Qwen3-4B & \textbf{29.0} & \textcolor{green!60!black}{+10.2} & \textbf{28.6} & \textcolor{green!60!black}{+3.6} & \textbf{78.2} & \textcolor{green!60!black}{+1.5} \\
Qwen3-8B & \textbf{26.9} & \textcolor{green!60!black}{+0.9} & \textbf{29.3} & \textcolor{green!60!black}{+3.5} & \textbf{80.7} & \textcolor{green!60!black}{+1.9} \\
Qwen2.5-7B & \textbf{16.8} & \textcolor{green!60!black}{+3.7} & \textbf{21.8} & \textcolor{green!60!black}{+6.8} & \textbf{73.4} & \textcolor{green!60!black}{+6.5} \\
Granite-4.1-8B & \textbf{33.5} & \textcolor{green!60!black}{+0.1} & \textbf{31.3} & \textcolor{green!60!black}{+6.4} & \textbf{77.9} & \textcolor{green!60!black}{+4.4} \\
\bottomrule
\end{tabular}
}
\end{table*}

\begin{figure*}[t]
\centering
\includegraphics[width=0.95\textwidth]{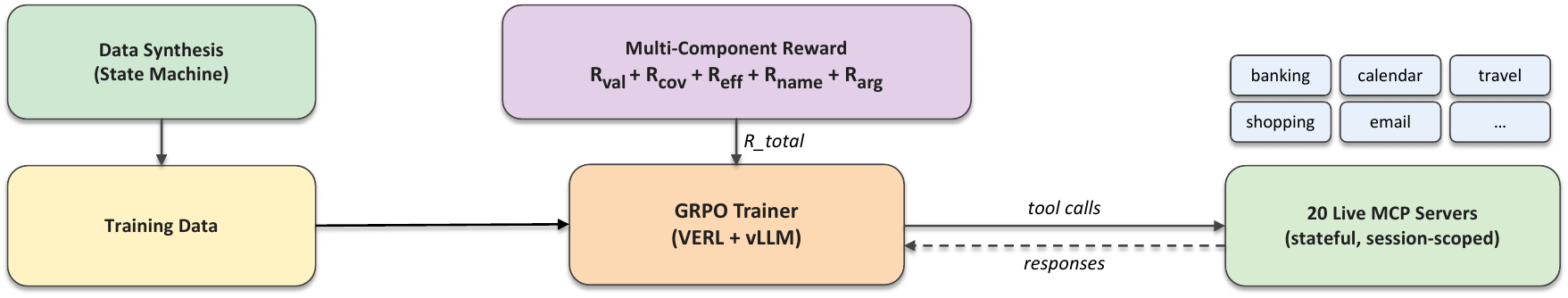}
\caption{System overview. An LLM teacher generates multi-turn trajectories against live MCP servers using auto-discovered dependency graphs. The GRPO trainer rolls out against the same servers and computes the five-component reward. Session-scoped state isolation ensures reproducibility.}
\label{fig:arch}
\end{figure*}

\noindent
We evaluate on BFCL Multi-Turn \citep{patil2024bfcl}, $\tau^2$-bench \citep{barres2025tau2}, and T-Eval \citep{chen2023teval}.
\ours improves all four models on all three benchmarks, using $\sim$13K training examples, roughly $8\times$ less than concurrent RL pipelines such as AgenticQwen \citep{agenticqwen2026}, and no judge model.
Figure~\ref{fig:arch} illustrates how the system works.

\section{Related Work}
\label{sec:related}


\paragraph{Data Synthesis and Supervised Fine-Tuning for Tool Use.}
A growing body of work scales synthetic data to train tool-using LLMs via supervised fine-tuning.
xLAM \citep{zhang2024xlam} releases a large corpus of verified single-turn function-calling traces, and APIGen-MT \citep{liu2025apigenmt} extends this to multi-turn trajectories with stronger verification.
Magnet \citep{yin2025magnet} synthesizes multi-turn traces by translating a tool-dependency graph into natural conversations and distilling them into smaller models.
ToolACE-MT \citep{zeng2026toolacemt} proposes non-autoregressive generation for agentic multi-turn interactions, and TOUCAN \citep{xu2025toucan} scales synthesis to 1.5M traces from real-world MCP servers.
Closer to our setting, \citet{crouse2026stateless} address the challenge of simulating complex multi-turn interactions when the execution backend is stateless.
These pipelines improve trajectory quality but typically feed a separate SFT stage and do not train against live execution feedback.
Our pipeline differs in three ways: (i) it is driven by a state machine that grounds queries in \emph{live} server state at generation time, (ii) it replay-validates every conversation against a freshly reset environment, and (iii) it produces $\sim$13K examples that feed directly into RL rollouts rather than an intermediate SFT corpus.

\paragraph{Reinforcement Learning for Tool Use.}
Reinforcement learning has become a popular way to elicit tool-use behavior without large SFT corpora.
ToolRL \citep{chen2025toolrl} showed that reward signals alone can teach tool-use capabilities, and ReTool \citep{feng2025retool} combined reasoning and tool use through strategic RL.
StepTool \citep{yu2024steptool} introduces step-grained rewards for multi-step tool chains, while Nemotron-Research-Tool-N1 \citep{zhang2025tooln1} explores reinforced reasoning for tool-using models.
Tool-R1 \citep{zhang2025toolr1} targets sample efficiency with compositional multi-step rewards, and MUA-RL \citep{zhao2025muarl} trains multi-turn policies against an LLM user simulator inside the RL loop.
Tool Zero \citep{zeng2025toolzero} trains tool-using agents via pure RL from scratch without SFT warm-up.
\citet{cheng2026toolorch} propose decomposing tool-use reward into per-call validity and sequence-level consistency components on a single cached-API domain; our validity term follows this graduated formulation but replaces AST comparison with live MCP schema validation and execution.
Concurrent work by \citet{agenticqwen2026} trains tool-use agents via GRPO with behavior-tree data flywheels ($\sim$100K examples) and LLM-as-judge rewards (Qwen3-235B).
Relative to this line, \ours differs along two axes: we train against \emph{live stateful} environments rather than cached or simulated APIs, and our reward is \emph{fully programmatic}---no external judge model---with an adaptive efficiency penalty that directly targets the verbosity incentive of pure-recall coverage rewards unaddressed in prior work.

\section{Method}
\label{sec:method}

We describe the three methodological components: the live MCP environment framework (\S\ref{sec:mcp}), the data synthesis pipeline (\S\ref{sec:data}), and the multi-component reward (\S\ref{sec:reward}).

\subsection{Live MCP Environment Framework}
\label{sec:mcp}

The Model Context Protocol \citep[MCP;][]{anthropic2024mcp} defines a standardized interface for LLMs to discover and invoke tools exposed by external servers.
We build a library of 20 MCP environment servers, each running as an independent subprocess communicating over stdio.
Each server exposes a domain-specific set of tools (10--40 per environment) with OpenAI-compatible function schemas.
The same set of environments is used for both data synthesis and RL training.

\paragraph{Architecture.}
An \texttt{MCPManager} orchestrates server lifecycle: starting, stopping, and resetting environments.
During RL training, a \texttt{MCPTool} wrapper integrates with the VERL framework \citep{sheng2024verl}, routing each model-generated tool call to the appropriate MCP server and returning the execution response.
Per-rollout state isolation is maintained via session-scoped state management: each rollout episode receives a unique session ID, ensuring that tool calls from one rollout cannot contaminate another's state.

\paragraph{Domain Coverage.}
The library covers 20 environments spanning finance, productivity, commerce, travel, social, IoT, developer tools, and knowledge management.
Table~\ref{tab:envs} lists the environments with their user-visible tool counts and characteristic state behavior.
Every environment is stateful---account balances change after transfers, calendar events persist, filesystem trees accumulate edits---providing realistic execution dynamics that static caches cannot capture.

\begin{table*}[t]
\centering
\small
\caption{The 20 MCP environments used for RL rollouts. All servers are stateful with session-scoped isolation; \textit{Tools} counts the user-visible tool surface exposed to the agent (helper/debug tools are hidden).}
\label{tab:envs}
\begin{tabular}{llcl}
\toprule
\textbf{Category} & \textbf{Environment} & \textbf{Tools} & \textbf{State / characteristic behavior} \\
\midrule
\multirow{3}{*}{Finance}
 & banking        & 17 & Accounts, transfers, balances; sensitive-param provenance \\
 & trading        & 13 & Portfolios, orders, market data; stateful order book \\
 & payments       & 10 & Invoices, refunds, webhooks; transactional state \\
\midrule
\multirow{3}{*}{Productivity}
 & calendar       & 17 & Events, invitees, recurring rules; time-dependent queries \\
 & email          & 17 & Inbox, drafts, threads, labels; append-only state \\
 & filesystem     & 40 & Files, dirs, permissions; deepest state (cd/ls/mv/cp chains) \\
\midrule
\multirow{3}{*}{Commerce}
 & shopping       & 23 & Catalog, cart, checkout; stateful cart \\
 & marketplace    & 13 & Listings, bids, seller ops; multi-entity state \\
 & retail\_chain  & 11 & Inventory across stores; location-scoped state \\
\midrule
\multirow{2}{*}{Travel}
 & travel\_booking & 20 & Flights, hotels, itineraries; booking state \\
 & maps           & 16 & Geocoding, routing, POIs; mostly stateless lookups \\
\midrule
\multirow{2}{*}{Social / Comm.}
 & social\_media  & 15 & Posts, follows, feeds; append + feed state \\
 & team\_chat     & 11 & Channels, messages, threads; append-only state \\
\midrule
IoT / Devices
 & iot\_devices   & 25 & Smart-home devices; sensor state + actuators \\
 & vehicle        & 20 & Ignition, climate, navigation; mode-gated actions \\
\midrule
\multirow{2}{*}{Knowledge / CRM}
 & crm            & 16 & Leads, contacts, deals; relational state \\
 & issue\_tracker & 20 & Issues, sprints, assignees; workflow transitions \\
 & budget         & 12 & Expenses, categories; accumulative state \\
\midrule
\multirow{2}{*}{Lifestyle}
 & food\_delivery & 17 & Menus, orders, tracking; order lifecycle state \\
 & video\_meeting & 10 & Meetings, participants, recordings; session state \\
\bottomrule
\end{tabular}
\end{table*}

\paragraph{Advantages over Static Caches.}
Compared to the cached-API approach of \citet{cheng2026toolorch}, live MCP execution offers:
(1)~\textit{Scalability}: adding a new domain requires only implementing an MCP server, not pre-computing a response cache;
(2)~\textit{Realism}: live execution captures state-dependent failures (e.g., insufficient balance, expired coupons) that static caches miss;
(3)~\textit{Determinism}: session-scoped state with deterministic reset ensures reproducible initial conditions.

\subsection{Data Synthesis Pipeline}
\label{sec:data}

\begin{figure*}[t]
\centering
\includegraphics[width=0.95\textwidth]{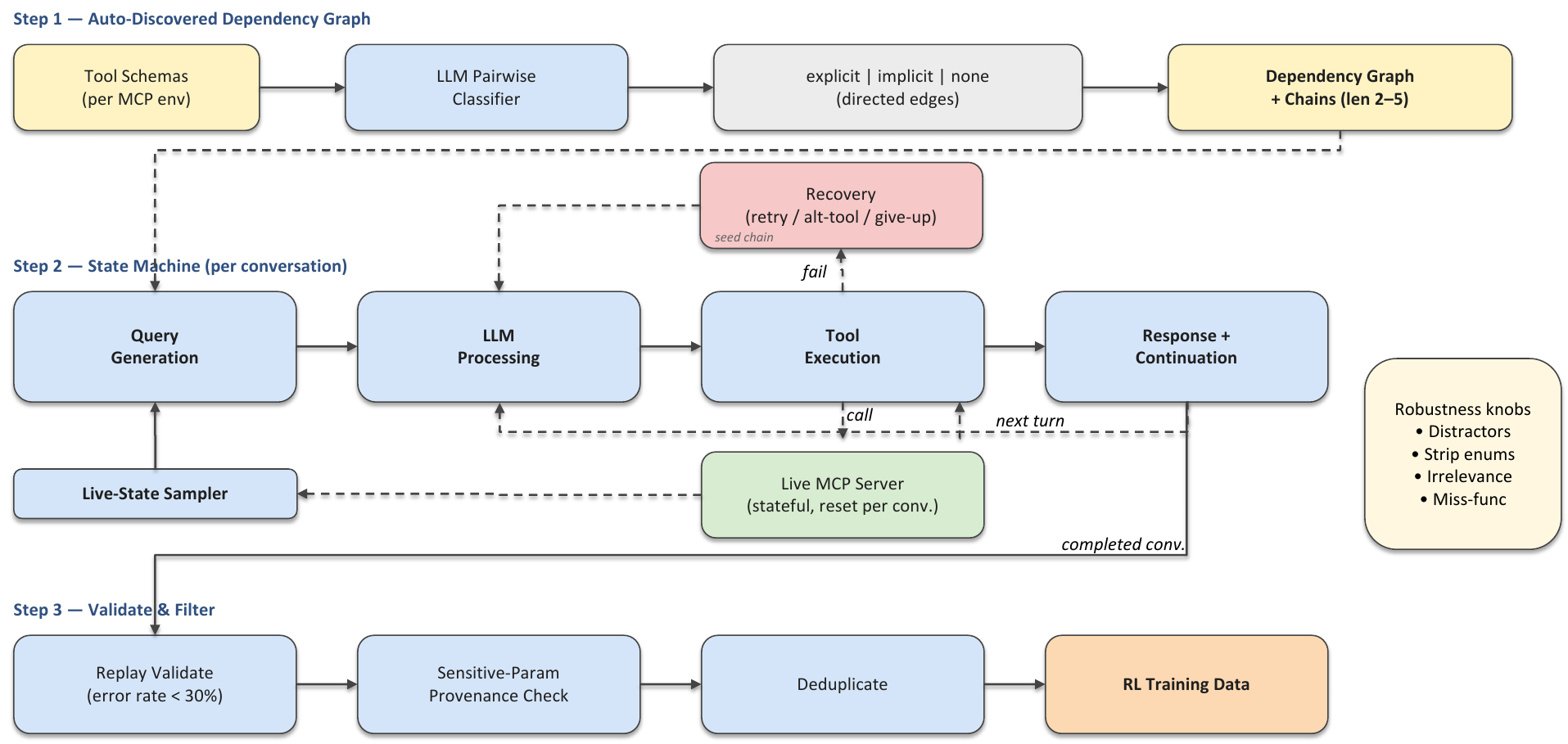}
\caption{State-machine-driven data synthesis. Per environment, we (1)~auto-discover a tool dependency graph, (2)~run a state machine that alternates query generation, tool execution, and continuation decisions against a live MCP server, and (3)~replay-validate each conversation before conversion to RL training data.}
\label{fig:datasyn}
\end{figure*}

We generate training data through a \emph{state-machine orchestrator} that drives a teacher LLM through multi-turn tool-use conversations against live MCP servers. Figure~\ref{fig:datasyn} illustrates the pipeline at a glance.

\paragraph{Step 1: Auto-Discovered Dependency Graph.}
For each environment, the pipeline probes all $\binom{n}{2}$ tool pairs and asks an LLM to classify each relationship as \textit{explicit} (output of tool A is a required input of tool B), \textit{implicit} (A must execute first to establish state), or \textit{none}. The resulting directed graph is hashed against the tool schema and cached, so the $O(n^2)$ cost is paid once per environment. From this graph we extract length-2 to length-5 tool chains that seed realistic multi-step queries.

\paragraph{Step 2: Live-State Sampling (Grounded Query Generation).}
A naive query generator would produce queries referencing fabricated entities---account \#12345, device \texttt{tv\_bedroom}, user \texttt{rachel@example.com}---that crash on execution because those entities do not exist in the server's state. We avoid this with a per-environment \textit{sampler} that probes the live MCP server \emph{before} query generation: it calls read-only discovery tools (e.g., \texttt{get\_unique\_values}, \texttt{search}, \texttt{list}) to enumerate real entities, filters them to those with sufficient supporting data (e.g., cities with at least 20 businesses, accounts with non-zero balance), and caches a compact ``sampling context'' per environment (refreshed every $k$ conversations). This context---real IDs, names, categories, and value ranges---is injected into the query-generation prompt along with an anti-hallucination instruction that constrains the LLM to use \emph{only} entities present in the context. The result is queries whose arguments exist in the server's actual state, making the chosen chain of tools executable end-to-end. For chain-seeded queries, the sampler additionally returns \textit{chain-aligned context}: a subset of entities most relevant to the specific tool chain being generated.

\paragraph{Step 3: State-Machine Orchestrator.}
Conversation generation is driven by an explicit finite state machine that tracks five groups of states---\textit{query}, \textit{turn}, \textit{tool-execution}, \textit{response}, and \textit{continuation}---with transitions triggered by the outcome of each LLM call or tool execution. A single turn proceeds as:
\begin{enumerate}
    \item \textsc{Query Generation}. A user query is sampled from the dependency-graph chains, grounded on the sampling context from Step~2, conditioned on a persona and reference date, and stratified by information level (60\% complete, 20\% missing-required, 20\% minimal).
    \item \textsc{LLM Processing}. The teacher LLM (Gemma-4-31B-it) is prompted with the query and tool schemas. A validator checks for well-formed tool calls; invalid responses trigger a fallback regeneration.
    \item \textsc{Tool Execution}. Tool calls are dispatched to the live MCP server. The outcome is classified as \textsc{success}, \textsc{partial-success}, or \textsc{failure}, and drives the next transition.
    \item \textsc{Recovery}. On failure, the state machine selects among retry-with-corrected-params, retry-with-alternative-tool, or graceful give-up with a fallback explanation.
    \item \textsc{Continuation}. A decision module samples whether to end the conversation, generate a follow-up, or ask a clarification, using a turn-decay schedule bounded by \texttt{min\_turns}$=2$ and \texttt{max\_turns}$=3$.
\end{enumerate}

\paragraph{Step 4: Robustness Knobs.}
Several probabilistic perturbations are applied during generation to broaden the training distribution:
(a)~\textit{Distractor injection} (40\%): 3--8 unrelated tools from other environments are added to the candidate set, forcing tool discrimination;
(b)~\textit{Enum stripping} (30\%): enumerated value lists are removed from parameter schemas, training the model to infer constraints from descriptions;
(c)~\textit{Irrelevance queries} (5\%): queries that no available tool can satisfy, training refusal behavior;
(d)~\textit{Missing-function}: a subset of tools is hidden after chain selection, producing abstention or clarification examples.

\paragraph{Step 5: Replay Validation and Deduplication.}
Each completed conversation is replayed against a freshly reset MCP environment. We count only schema-level and execution errors (not empty-result responses) and discard conversations with error rate above 30\%. A provenance check ensures that sensitive parameters (passwords, tokens) appear only when traceable to prior user turns or tool outputs. Surviving conversations are deduplicated by Jaccard similarity on tool-call sequences (threshold 0.70).

\paragraph{Training Data.}
The final training set comprises 13{,}517 examples drawn from three sources: 10{,}895 multi-turn MCP conversations across 20 environments; 1{,}500 clarification trajectories generated by the missing-function variant of Step 3; and 1{,}122 externally-sourced abstention examples ($G=\emptyset$)---806 from When2Call \citep{ross2025when2call} (no-tool-call scenarios) and 316 from xLAM-Irrelevance \citep{madeagents2024xlamirrelevance} (irrelevant queries). All models are trained in a single stage for 350 GRPO steps.

\subsection{Multi-Component Reward}
\label{sec:reward}

Our reward decomposes into five components:
\textbf{validity} (is each call structurally correct and executable?),
\textbf{coverage} (are all required steps completed in dependency order?),
\textbf{efficiency} (was it done concisely?),
\textbf{tool-name guidance} (did the model select the right tools?), and
\textbf{argument-value matching} (do the chosen argument \emph{values} agree with ground truth, not just the argument keys?).
The first two address natural desiderata for any tool-calling system. The efficiency penalty---specifically its complexity-scaled, adaptive budget---and the tool-name bonus are our two most load-bearing reward-side contributions: an ablation (\S\ref{sec:reward_ablation}) shows each is responsible for a $\sim$9-point aggregate drop across the three benchmarks when removed. Argument-value matching is a smaller-weight signal that disproportionately helps multi-turn dialog ($\tau^2$).

\paragraph{Validity ($\rvalidity$).}
Each tool call is scored on three graduated levels, equally weighted: (1)~the function name exists in the candidate schema; (2)~all required parameters are present with compatible JSON types; (3)~live execution against the MCP server succeeds without error.
The validity reward averages these per-call scores across all $N$ model calls.
This graduated formulation follows \citet{cheng2026toolorch}, though we replace their AST-based check with live MCP schema validation and execution.
It provides partial credit---a call with the correct name but wrong parameters receives $\sim$0.33; correct structure but failed execution receives $\sim$0.66---giving the model fine-grained learning signal even on partially-correct attempts.

\paragraph{Coverage ($\rcoverage$).}
Measures whether the model's output covers all ground-truth workflow steps in the correct dependency order:
\begin{equation}
\rcoverage = \frac{1}{|G|} \sum_{g \in G} m(g) \cdot o(g)
\end{equation}
where $m(g) = \mathbb{1}[g \text{ matched}]$ and $o(g) = \prod_{(j,g) \in E} \mathbb{1}[\mu(j) < \mu(g)]$ enforces dependency order; $G$ is the set of ground-truth steps, $E$ the dependency edges, and $\mu(\cdot)$ maps steps to their matched position in the model's output sequence.
A model call matches a GT step only if it includes all of the GT call's \textbf{argument keys}---preventing superficial matches where the model calls the right function with entirely wrong parameters.

\paragraph{Adaptive Efficiency ($\refficiency$).}
Our key contribution. Coverage reward (pure recall) incentivizes verbosity: more tool calls increase the probability of matching GT steps, with no penalty for excess.
We introduce an additive penalty with a \textit{complexity-adaptive budget}:
\begin{align}
\refficiency &= -\alpha \cdot \frac{\max(0,\; n_\text{calls} - B)}{B} \label{eq:reff} \\
B &= n_\text{gt} + \lceil n_\text{gt} \cdot \beta \rceil \nonumber
\end{align}
where $n_\text{calls}$ is the number of model tool calls, $n_\text{gt}$ is the total number of ground-truth tool calls (summed across all steps, including parallel calls within a single step), $\alpha$ controls penalty strength, and $\beta$ controls the slack budget.
Counting calls rather than steps is what makes the penalty consistent: a step with three parallel ground-truth calls budgets three calls, not one, so a model that emits the correct three is not penalized as if it had three excess calls.

The adaptive budget $B$ is critical: a flat penalty ($B = n_\text{gt}$) over-penalizes complex tasks where exploration or information-gathering calls are legitimate.
With $\beta = 0.5$, a 4-call task has budget $B = 6$, allowing 2 extra calls before any penalty applies.
Key properties:
\begin{itemize}
    \item \textit{Complexity-adaptive}: Harder tasks (more GT steps) receive proportionally more slack.
    \item \textit{Asymmetric}: Only penalizes \emph{excess} calls (not missing calls, which are already captured by $\rcoverage$).
    \item \textit{GRPO-compatible}: Under group-relative advantage normalization, the penalty naturally scales with batch difficulty.
\end{itemize}

\paragraph{Tool Name Bonus ($R_\text{name}$).}
We observed in pilot training runs that tool-name selection is a primary early-training bottleneck: validity ($\rvalidity$) saturates well before the model reliably picks the right tools, leaving the coverage signal too sparse to drive consistent learning.
We add an explicit reward that credits the model for calling tools whose names appear in the ground-truth set:
\begin{equation}
R_\text{name} = \frac{|\{c \in \hat{C} : c.\text{name} \in \text{GT\_names}\}|}{|\hat{C}|}
\end{equation}
where $\hat{C}$ is the set of model tool calls and GT\_names is the set of tool names appearing in the ground truth.
This is independent of ordering or argument matching---it purely rewards tool \textit{selection} quality.

\paragraph{Argument-Value Matching ($R_\text{arg}$).}
Coverage matches a model call to a GT step using the function name and the set of \emph{argument keys}---it does not compare the actual argument values, so a model that fills the correct slots with the wrong values still receives full coverage credit. To close this gap we add a value-matching bonus: for each pair of GT and model calls already aligned by name and key set, we score the fraction of argument values that match GT, and average across aligned pairs:
\begin{equation}
R_\text{arg} = \frac{1}{|M|} \sum_{(c, g) \in M} \frac{|\{k : c.\text{args}[k] = g.\text{args}[k]\}|}{|g.\text{args}|}
\end{equation}
where $M$ is the set of (model call, GT call) pairs already aligned by Coverage. This signal is small in magnitude but consistently lifts BFCL Multi-Turn, $\tau^2$-bench, and T-Eval at $w_\text{arg} = 0.1$; we did not observe similar gains at much higher or much lower weights, suggesting $0.1$ is a sweet spot that nudges value choice without overshadowing the main reward terms.

\paragraph{Abstention Reward.}
For abstention examples ($G = \emptyset$): $R = 1.0$ if zero tool calls (correct abstention), $R = 0.0$ otherwise.

\paragraph{Combined Reward.}
\begin{align}
\rtotal = &\; w_\text{val}\, \rvalidity + w_\text{cov}\, \rcoverage \nonumber \\
&+ w_\text{eff}\, \refficiency + w_\text{name}\, R_\text{name} \nonumber \\
&+ w_\text{arg}\, R_\text{arg}
\label{eq:rtotal}
\end{align}
with $w_\text{val} = 0.5$, $w_\text{cov} = 0.5$, $w_\text{eff} = 0.15$, $w_\text{name} = 0.2$, $w_\text{arg} = 0.1$, $\alpha = 0.5$, and $\beta = 0.5$ throughout all experiments.
The auxiliary weights ($w_\text{eff}$, $w_\text{name}$, $w_\text{arg}$) are kept smaller than the main components ($w_\text{val}$, $w_\text{cov}$) to avoid suppressing exploration early in training.


\section{Experiments}
\label{sec:experiments}

\subsection{Setup}

\paragraph{Base Models.}
We evaluate on four instruction-tuned models spanning two architecture families:
Qwen3-4B-Instruct\footnote{\url{https://huggingface.co/Qwen/Qwen3-4B-Instruct-2507}}, Qwen3-8B\footnote{\url{https://huggingface.co/Qwen/Qwen3-8B}} \citep{qwen3}, Qwen2.5-7B-Instruct\footnote{\url{https://huggingface.co/Qwen/Qwen2.5-7B-Instruct}} \citep{qwen25}, and Granite-4.1-8B-Instruct\footnote{\url{https://huggingface.co/ibm-granite/granite-4.1-8b-instruct}} \citep{granite41}.
This covers 4B--8B parameters across Qwen and Granite architectures, testing whether a single reward configuration generalizes across model families.

\paragraph{Training.}
We train with Group Relative Policy Optimization \citep[GRPO;][]{shao2024deepseekmath} via the VERL framework \citep{sheng2024verl} on 8$\times$H100 GPUs.
Key hyperparameters: batch size 16, rollout group size 16, tensor parallelism 2.
Maximum prompt length is 12,384 tokens; maximum response length is 16,384 tokens.
Multi-turn rollouts are capped at 5 user turns and 10 assistant turns.
RL rollouts use 20 live MCP environments (see \S\ref{sec:mcp}).
Training uses 13{,}517 samples (10{,}895 multi-turn MCP conversations + 1{,}500 clarification + 806 When2Call + 316 xLAM-Irrelevance) for 350 GRPO steps in a single stage.
KL coefficient is $0.01$ for all models. Learning rate is the only per-family knob: Qwen3-4B uses $1\times10^{-6}$, Qwen3-8B uses $5\times10^{-8}$, Qwen2.5-7B uses $3\times10^{-7}$, and Granite-4.1-8B uses $3\times10^{-7}$. The Qwen2.5 and Granite families are more sensitive to learning rate than Qwen3-4B---at $1\times10^{-6}$ both regress sharply on multi-turn benchmarks (see \S\ref{sec:experiments} discussion); we selected each model's LR from a small sweep over $\{1\times10^{-6},\, 3\times10^{-7},\, 5\times10^{-8}\}$.
All models use identical reward configuration ($w_\text{val}=0.5$, $w_\text{cov}=0.5$, $w_\text{eff}=0.15$, $w_\text{name}=0.2$, $w_\text{arg}=0.1$, $\alpha=0.5$, $\beta=0.5$).

\paragraph{Benchmarks.}
We evaluate on three benchmarks:
\begin{itemize}
    \item \textbf{BFCL Multi-Turn} \citep{zhong2025complexfuncbench}: Evaluates multi-step function calling with subcategories for base multi-turn, missing function, missing parameter, and long context scenarios. We report Overall MT accuracy and all subcategories.
    \item \textbf{$\tau^2$-bench} \citep{barres2025tau2}: Evaluates conversational agents in a dual-control environment where both agent and user can modify shared state across airline, retail, and telecom domains. An LLM user simulator drives multi-turn conversations where the agent must call tools to resolve requests. We report per-domain and average task completion rates.
    \item \textbf{T-Eval} \citep{chen2023teval}: Evaluates tool-use capabilities across six dimensions: instruction following, planning, reasoning, retrieval, understanding, and review. We report scores averaged over JSON and string prompt formats.
\end{itemize}
For $\tau^2$-bench, which requires an LLM user simulator to drive multi-turn interactions, we use \texttt{Qwen3-VL-235B-A22B-Instruct}\footnote{\url{https://huggingface.co/Qwen/Qwen3-VL-235B-A22B-Instruct}} as the simulator across all runs to keep evaluations directly comparable.

\subsection{Main Results}

Table~\ref{tab:main_results} presents our main results comparing base models against our method on BFCL Multi-Turn and $\tau^2$-bench.

\begin{table*}[t]
\centering
\caption{Main results across model families. All models trained for 350 GRPO steps with \ours{}.}
\label{tab:main_results}
\resizebox{0.95\textwidth}{!}{%
\begin{tabular}{ll|ccccc|cccc}
\toprule
\multirow{2}{*}{\textbf{Model}} & \multirow{2}{*}{\textbf{Method}} & \multicolumn{5}{c|}{\textbf{BFCL Multi-Turn}} & \multicolumn{4}{c}{\textbf{$\tau^2$-bench}} \\
 & & Base & Miss Param & Miss Func & Long Ctx & Overall & Airline & Retail & Telecom & Avg \\
\midrule
Qwen3-4B & Base & 23.5 & 14.5 & 14.0 & 23.0 & 18.8 & 26.8 & 35.1 & 13.2 & 25.0 \\
 & \ours{} & \textbf{35.5} & \textbf{18.5} & \textbf{29.0} & \textbf{33.0} & \textbf{29.0} & \textbf{28.0} & \textbf{42.1} & \textbf{15.8} & \textbf{28.6} \\
 & $\Delta$ & \textcolor{green!60!black}{+12.0} & \textcolor{green!60!black}{+4.0} & \textcolor{green!60!black}{+15.0} & \textcolor{green!60!black}{+10.0} & \textcolor{green!60!black}{+10.2} & \textcolor{green!60!black}{+1.2} & \textcolor{green!60!black}{+7.0} & \textcolor{green!60!black}{+2.6} & \textcolor{green!60!black}{+3.6} \\
\midrule
Qwen3-8B & Base & 30.5 & 19.5 & 31.5 & \textbf{22.5} & 26.0 & 24.0 & 28.9 & \textbf{24.6} & 25.8 \\
 & \ours{} & \textbf{34.0} & \textbf{21.5} & \textbf{32.0} & 20.0 & \textbf{26.9} & \textbf{30.0} & \textbf{34.2} & 23.7 & \textbf{29.3} \\
 & $\Delta$ & \textcolor{green!60!black}{+3.5} & \textcolor{green!60!black}{+2.0} & \textcolor{green!60!black}{+0.5} & \textcolor{red}{-2.5} & \textcolor{green!60!black}{+0.9} & \textcolor{green!60!black}{+6.0} & \textcolor{green!60!black}{+5.3} & \textcolor{red}{-0.9} & \textcolor{green!60!black}{+3.5} \\
\midrule
Qwen2.5-7B & Base & 17.0 & 10.0 & 14.5 & 11.0 & 13.1 & 16.0 & 12.3 & 16.7 & 15.0 \\
 & \ours{} & \textbf{22.5} & \textbf{13.5} & \textbf{18.0} & \textbf{13.0} & \textbf{16.8} & \textbf{26.0} & \textbf{16.7} & \textbf{22.8} & \textbf{21.8} \\
 & $\Delta$ & \textcolor{green!60!black}{+5.5} & \textcolor{green!60!black}{+3.5} & \textcolor{green!60!black}{+3.5} & \textcolor{green!60!black}{+2.0} & \textcolor{green!60!black}{+3.7} & \textcolor{green!60!black}{+10.0} & \textcolor{green!60!black}{+4.4} & \textcolor{green!60!black}{+6.1} & \textcolor{green!60!black}{+6.8} \\
\midrule
Granite-4.1-8B & Base & \textbf{44.5} & \textbf{26.0} & 23.5 & \textbf{39.5} & 33.4 & 8.0 & \textbf{49.1} & 17.5 & 24.9 \\
 & \ours{} & 43.0 & 24.0 & \textbf{30.0} & 37.0 & \textbf{33.5} & \textbf{22.0} & 47.4 & \textbf{24.6} & \textbf{31.3} \\
 & $\Delta$ & \textcolor{red}{-1.5} & \textcolor{red}{-2.0} & \textcolor{green!60!black}{+6.5} & \textcolor{red}{-2.5} & \textcolor{green!60!black}{+0.1} & \textcolor{green!60!black}{+14.0} & \textcolor{red}{-1.8} & \textcolor{green!60!black}{+7.0} & \textcolor{green!60!black}{+6.4} \\
\bottomrule
\end{tabular}}
\end{table*}

\paragraph{BFCL Multi-Turn.}
All four models improve on BFCL MT Overall: Qwen3-4B (+10.2), Qwen2.5-7B (+3.7), Qwen3-8B (+0.9), and Granite-4.1-8B (+0.1, essentially flat).
The MT Base subcategory shows the strongest gain on Qwen3-4B (+12.0), indicating substantially improved multi-step orchestration; Granite-4.1-8B's near-zero overall change masks a +6.5 gain on Multi-Turn Miss-Function offset by smaller regressions on Base ($-1.5$) and Long-Context ($-2.5$).

\paragraph{$\tau^2$-bench.}
All four models improve: Qwen2.5-7B (+6.8), Granite-4.1-8B (+6.4), Qwen3-4B (+3.6), and Qwen3-8B (+3.5).
Granite-4.1-8B's gain is concentrated on the airline domain (+14.0, from 8.0 to 22.0) and telecom (+7.0); Qwen2.5-7B's gain is the most uniform across the three domains, with all three improving by $+4$ pts or more.
The earlier divergence pattern we observed for Qwen3-8B at the launcher's default learning rate ($1\times10^{-6}$) disappears at the lower $5\times10^{-8}$ used here---the model stabilizes and posts a TAU2 gain comparable to Qwen3-4B.

\subsection{T-Eval Results}

Table~\ref{tab:teval} reports T-Eval scores across all four models.

\begin{table*}[t]
\centering
\caption{T-Eval results (0--100). Scores averaged over JSON and string prompt formats. All models trained for 350 steps with \ours{}.}
\label{tab:teval}
\resizebox{0.7\textwidth}{!}{%
\begin{tabular}{llrrrrrrr}
\toprule
Model & Method & Instruct & Plan & Reason & Retrieve & Understand & Review & \textbf{Overall} \\
\midrule
Qwen3-4B & Base & 95.5 & 66.4 & 62.8 & 87.5 & 77.3 & \textbf{70.8} & 76.7 \\
Qwen3-4B & \ours{} & \textbf{98.5} & \textbf{68.7} & \textbf{63.7} & \textbf{90.6} & \textbf{78.7} & 68.8 & \textbf{78.2} \\
 & $\Delta$ & \textcolor{green!60!black}{+3.0} & \textcolor{green!60!black}{+2.3} & \textcolor{green!60!black}{+1.0} & \textcolor{green!60!black}{+3.1} & \textcolor{green!60!black}{+1.4} & \textcolor{red}{-2.1} & \textcolor{green!60!black}{+1.5} \\
\midrule
Qwen3-8B & Base & \textbf{98.8} & 64.7 & \textbf{64.2} & 88.0 & 77.8 & 79.2 & 78.8 \\
Qwen3-8B & \ours{} & 94.8 & \textbf{65.2} & 63.0 & \textbf{92.7} & \textbf{81.8} & \textbf{86.5} & \textbf{80.7} \\
 & $\Delta$ & \textcolor{red}{-4.1} & \textcolor{green!60!black}{+0.4} & \textcolor{red}{-1.2} & \textcolor{green!60!black}{+4.7} & \textcolor{green!60!black}{+4.0} & \textcolor{green!60!black}{+7.3} & \textcolor{green!60!black}{+1.9} \\
\midrule
Qwen2.5-7B & Base & 73.4 & 54.1 & 57.7 & 79.2 & 62.2 & 75.0 & 66.9 \\
Qwen2.5-7B & \ours{} & \textbf{81.8} & \textbf{60.4} & \textbf{63.1} & \textbf{88.0} & \textbf{69.7} & \textbf{77.1} & \textbf{73.4} \\
 & $\Delta$ & \textcolor{green!60!black}{+8.4} & \textcolor{green!60!black}{+6.3} & \textcolor{green!60!black}{+5.5} & \textcolor{green!60!black}{+8.9} & \textcolor{green!60!black}{+7.6} & \textcolor{green!60!black}{+2.1} & \textcolor{green!60!black}{+6.5} \\
\midrule
Granite-4.1-8B & Base & 93.5 & 68.2 & 63.4 & 87.0 & 74.0 & 55.2 & 73.5 \\
Granite-4.1-8B & \ours{} & \textbf{94.7} & \textbf{69.7} & \textbf{63.7} & \textbf{91.7} & \textbf{76.9} & \textbf{70.8} & \textbf{77.9} \\
 & $\Delta$ & \textcolor{green!60!black}{+1.1} & \textcolor{green!60!black}{+1.5} & \textcolor{green!60!black}{+0.4} & \textcolor{green!60!black}{+4.7} & \textcolor{green!60!black}{+2.9} & \textcolor{green!60!black}{+15.6} & \textcolor{green!60!black}{+4.4} \\
\bottomrule
\end{tabular}}
\end{table*}

\paragraph{Analysis.}
All four models improve on T-Eval overall: Qwen2.5-7B (+6.5), Granite-4.1-8B (+4.4), Qwen3-8B (+1.9), and Qwen3-4B (+1.5).
Qwen2.5-7B shows the strongest gains across five of six dimensions; Granite-4.1-8B posts the single largest improvement on review (+15.6).
The retrieve and plan dimensions benefit most consistently across models, as these directly exercise the multi-step tool-dependency skills our reward targets.

\subsection{Training Dynamics}

\begin{figure}[t]
\centering
\includegraphics[width=\columnwidth]{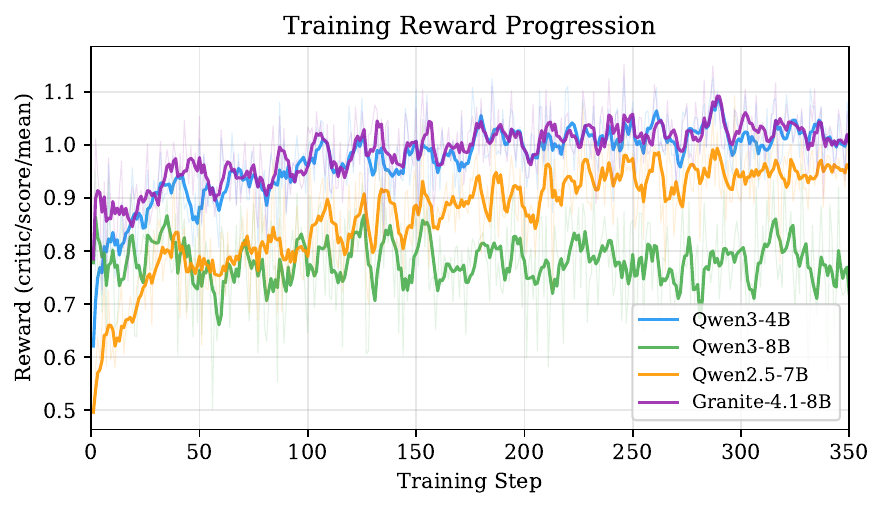}
\caption{Training reward (critic/score/mean) over 350 GRPO steps. All models converge smoothly without instability.}
\label{fig:training_curves}
\end{figure}

Figure~\ref{fig:training_curves} shows the training reward progression over 350 steps.
Three of four models exhibit rapid initial improvement (steps 0--50) followed by gradual refinement, confirming that the reward provides strong learning signal from the start; Qwen3-8B is comparatively flat throughout.

\begin{figure*}[t]
\centering
\includegraphics[width=\textwidth]{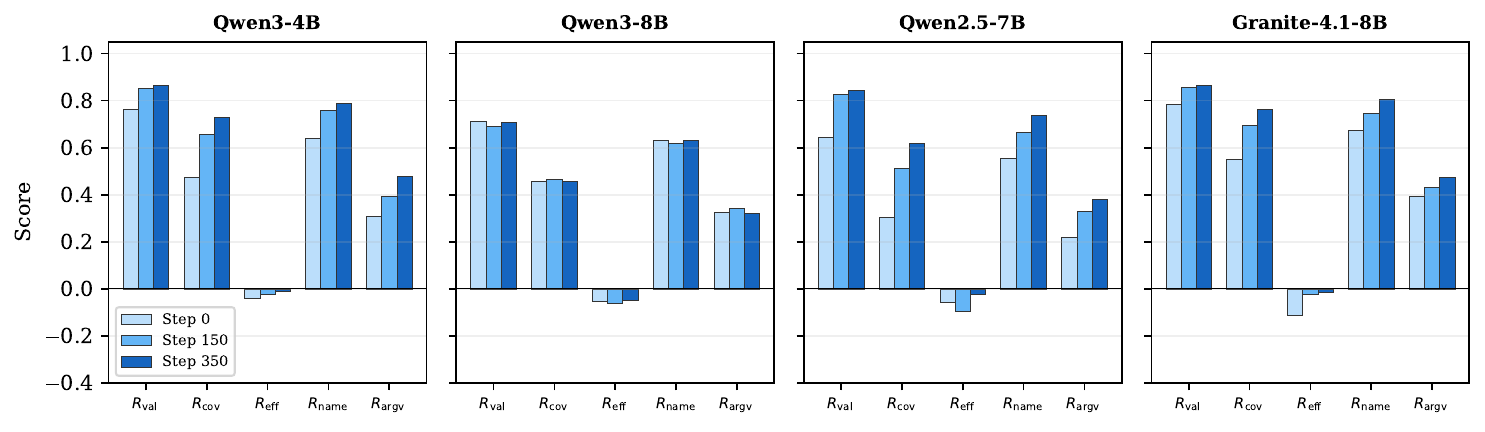}
\caption{Reward component progression at steps 0, 150, and 350 across all four models. Coverage ($R_\text{cov}$) shows the largest and most consistent improvement; validity ($R_\text{val}$) saturates early.}
\label{fig:components}
\end{figure*}

Figure~\ref{fig:components} decomposes the reward into its five components at steps 0, 150, and 350 across all models.
Across Qwen3-4B, Qwen2.5-7B, and Granite-4.1-8B, the four trainable components move in the expected directions: $R_\text{cov}$ shows the largest gains, $R_\text{val}$ saturates early (most improvement by step 150), $R_\text{eff}$ shrinks toward zero as models learn conciseness, and $R_\text{name}$ improves steadily.
Qwen3-8B is the exception: every component is essentially flat throughout training, consistent with the conservative learning rate required to stabilize this base model and with its smaller eval-time gains.

\subsection{Reward-Component Ablation}
\label{sec:reward_ablation}

To verify that each reward term contributes, we re-train Qwen3-4B with the paper recipe (350 GRPO steps) under six variants: each of the five reward components zeroed in turn, plus a flat-budget variant that disables the adaptive efficiency target ($\beta\!=\!0$).
Table~\ref{tab:reward_ablation} reports BFCL Multi-Turn, $\tau^2$-bench average, and T-Eval Overall against the full reward.

\begin{table}[t]
\centering
\caption{Reward-component ablation on Qwen3-4B with the paper recipe (350 GRPO steps). Each row drops one term from the combined reward; the last row keeps all terms but flattens the adaptive efficiency budget ($\beta\!=\!0$). $\Delta$ vs.\ the full reward is shown in parentheses. TAU2 is reported on a 0--100 scale. Code-side names map as $\rvalidity\!\equiv\!R_\text{atomic}$, $\rcoverage\!\equiv\!R_\text{orch}$, $\refficiency\!\equiv\!R_\text{eff}$.}
\label{tab:reward_ablation}
\resizebox{\columnwidth}{!}{%
\begin{tabular}{l|ccc}
\toprule
\textbf{Variant} & \textbf{BFCL MT} & \textbf{TAU2} & \textbf{T-Eval} \\
\midrule
Full reward (\ours) & \textbf{29.0} & \textbf{28.6} & \textbf{78.2} \\
\midrule
\;\;\textminus $\rvalidity$ & 23.8 {\scriptsize \textcolor{red}{-5.2}} & 30.7 {\scriptsize \textcolor{green!60!black}{+2.1}} & 74.8 {\scriptsize \textcolor{red}{-3.4}} \\
\;\;\textminus $\rcoverage$ & 26.6 {\scriptsize \textcolor{red}{-2.4}} & 29.0 {\scriptsize \textcolor{green!60!black}{+0.4}} & 76.7 {\scriptsize \textcolor{red}{-1.5}} \\
\;\;\textminus $\refficiency$ & 25.0 {\scriptsize \textcolor{red}{-4.0}} & 28.5 {\scriptsize \textcolor{red}{-0.1}} & 76.5 {\scriptsize \textcolor{red}{-1.6}} \\
\;\;\textminus $R_\text{name}$ & 23.0 {\scriptsize \textcolor{red}{-6.0}} & 27.8 {\scriptsize \textcolor{red}{-0.8}} & 75.8 {\scriptsize \textcolor{red}{-2.4}} \\
\;\;\textminus $R_\text{arg}$ & 30.6 {\scriptsize \textcolor{green!60!black}{+1.6}} & 24.5 {\scriptsize \textcolor{red}{-4.2}} & 77.6 {\scriptsize \textcolor{red}{-0.6}} \\
$\refficiency$ flat ($\beta\!=\!0$) & 22.4 {\scriptsize \textcolor{red}{-6.6}} & 25.5 {\scriptsize \textcolor{red}{-3.1}} & 78.6 {\scriptsize \textcolor{green!60!black}{+0.4}} \\
\bottomrule
\end{tabular}}
\end{table}

Every component carries weight on at least one benchmark.
Removing $R_\text{name}$ produces the largest broad-spectrum drop (BFCL MT $-6.0$, T-Eval $-2.4$); removing $R_\text{arg}$ collapses $\tau^2$ ($-4.2$ pts) while paradoxically lifting MT (the model relearns to fire calls aggressively without the argument-quality penalty).
Dropping $\rvalidity$ or $\rcoverage$ each costs MT $-2$ to $-5$ pts together with T-Eval drops, confirming both terms shape generalisation beyond their specific signals.
The strongest single result is the $\beta\!=\!0$ row: flattening the adaptive budget in $\refficiency$ loses MT $-6.6$ and $\tau^2$ $-3.1$, isolating the per-task ground-truth-call target as the most load-bearing piece of the efficiency formulation.

\subsection{Comparison to Supervised Fine-Tuning}
\label{sec:sft_comparison}

To isolate the contribution of the reward signal, we train an SFT baseline on the same training data (3 epochs, ms-swift) for each model and evaluate alongside the base model and \ours{}.
Table~\ref{tab:sft_comparison} reports BFCL Multi-Turn, $\tau^2$-bench, and T-Eval Overall.

\begin{table}[t]
\centering
\caption{Comparing \ours{} (RL with our reward) against SFT on the same training data and the base model. SFT is trained for 3 epochs with ms-swift; \ours{} runs 350 GRPO steps. \ours{} dominates SFT on all three benchmarks for 3/4 models, and ties or wins on the fourth. Best result per row in bold. $\tau^2$ on a 0--100 scale.}
\label{tab:sft_comparison}
\resizebox{\columnwidth}{!}{%
\begin{tabular}{ll|ccc}
\toprule
\textbf{Model} & \textbf{Method} & \textbf{BFCL MT} & \textbf{$\tau^2$} & \textbf{T-Eval} \\
\midrule
Qwen3-4B & Base & 18.8 & 25.0 & 76.7 \\
 & SFT & 26.2 & 19.6 & 74.7 \\
 & \ours{} & \textbf{29.0} & \textbf{28.6} & \textbf{78.2} \\
\midrule
Qwen3-8B & Base & 26.0 & 25.8 & 78.8 \\
 & SFT & 22.4 & 11.5 & 76.3 \\
 & \ours{} & \textbf{26.9} & \textbf{29.3} & \textbf{80.7} \\
\midrule
Qwen2.5-7B & Base & 13.1 & 15.0 & 66.9 \\
 & SFT & \textbf{22.5} & 15.2 & 72.2 \\
 & \ours{} & 16.8 & \textbf{21.8} & \textbf{73.4} \\
\midrule
Granite-4.1-8B & Base & 33.4 & 24.9 & 73.5 \\
 & SFT & 29.0 & 19.6 & 75.2 \\
 & \ours{} & \textbf{33.5} & \textbf{31.3} & \textbf{77.9} \\
\bottomrule
\end{tabular}}
\end{table}

\ours{} dominates SFT on 11 of 12 cells (4 models $\times$ 3 benchmarks), only ceding BFCL MT on Qwen2.5-7B.
The most striking pattern is on $\tau^2$-bench: SFT regresses below the base model on 3/4 models (Qwen3-4B $-5.4$, Qwen3-8B $-14.3$, Granite-4.1-8B $-5.3$ pts), suggesting that imitating the synthetic trajectories without execution-grounded reward damages the multi-turn agentic behaviour the base model already had.
\ours{} never regresses on $\tau^2$ relative to base, and improves it by $+3.5$ to $+6.8$ across all four models.
The same effect is visible but milder on T-Eval, where SFT trails \ours{} by $1.2$--$4.4$ pts.
This indicates that the data alone is insufficient: the reward signal is what converts the trajectories into reliable multi-step orchestration.

\subsection{Teacher Ablation}
\label{sec:teacher_ablation}

The orchestrator and clarification pipelines (\S\ref{sec:data}) use a teacher LLM to drive query generation and assistant turns.
To check whether our results depend on a particular teacher, we regenerate the full training mixture using Qwen3-32B as the teacher in place of Gemma-4-31B-it, keeping every other knob fixed (same prompts, same $\binom{n}{2}$ dependency-graph cache, same filtering).
We then train Qwen3-4B and Qwen3-8B with the paper recipe (350 steps, $R_\text{arg}\!=\!0.1$, family-specific LR).

\begin{table}[t!]
\centering
\caption{Teacher ablation. We re-run the data-synthesis pipeline (orchestrator + clarification) with Qwen3-32B as the teacher in place of Gemma-4-31B-it, then train Qwen3-4B and Qwen3-8B with the paper recipe (350 GRPO steps). Results are within noise across both teachers on all three benchmarks, and both improve over the base model. Best result per row in bold; $\tau^2$ on a 0--100 scale.}
\label{tab:teacher_ablation}
\resizebox{\columnwidth}{!}{%
\begin{tabular}{ll|ccc}
\toprule
\textbf{Model} & \textbf{Setting} & \textbf{BFCL MT} & \textbf{$\tau^2$} & \textbf{T-Eval} \\
\midrule
Qwen3-4B & Base  & 18.8 & 25.0 & 76.7 \\
 & Gemma-31B & \textbf{29.0} & 28.6 & \textbf{78.2} \\
 & Qwen3-32B & 22.6 & \textbf{29.0} & 76.8 \\
\midrule
Qwen3-8B & Base  & 26.0 & 25.8 & 78.8 \\
 & Gemma-31B & 26.9 & 29.3 & \textbf{80.7} \\
 & Qwen3-32B & \textbf{27.1} & \textbf{29.7} & 79.8 \\
\bottomrule
\end{tabular}}
\end{table}

Results in Table~\ref{tab:teacher_ablation} are within noise on all three benchmarks: $\tau^2$-bench scores are essentially identical across teachers ($\Delta\!<\!1$ pt for both models) and T-Eval differs by at most $1.4$ pts.
BFCL Multi-Turn shows a Qwen3-4B regression with the Qwen3-32B teacher ($-6.4$ pts) but no such effect on Qwen3-8B ($+0.2$).
This robustness to teacher choice suggests the method's gains stem from the reward formulation and the live-execution training loop rather than a particular teacher's writing style.




\section{Conclusion}
\label{sec:conclusion}

We presented \oursexpanded, which couples live stateful MCP environments, state-grounded multi-turn data synthesis, and a programmatic multi-component RL reward into a single training loop---no LLM-as-judge, no manual annotation.
Two reward-side components carry the most weight: an adaptive efficiency penalty $\refficiency$ with a complexity-scaled call budget, which counters the verbosity incentive of pure-recall coverage rewards, and a tool-name signal $R_\text{name}$; an ablation (\S\ref{sec:reward_ablation}) shows each is responsible for a $\sim$9-point aggregate drop across the three benchmarks when removed. Validity, coverage, and an argument-value matching signal (the latter disproportionately helping multi-turn dialog) round out the five-component reward.
With a single reward configuration and only learning rate tuned per family, four models from two architecture families improve on BFCL Multi-Turn, $\tau^2$-bench, and T-Eval, supporting our central claim: a compact, fully programmatic reward coupled with grounded data synthesis can drive consistent multi-step tool-orchestration gains without large-scale data pipelines or expensive judge models.



\bibliographystyle{plainnat}
\bibliography{references}

\end{document}